\pdfoutput=1
\documentclass{article}

\usepackage{microtype}
\usepackage{graphicx}
\usepackage{subcaption}
\usepackage{booktabs}
\usepackage{multirow}
\usepackage{tikz}
\usetikzlibrary{arrows.meta,positioning,fit,backgrounds,calc}

\usepackage{hyperref}

\usepackage[preprint]{icml2026}

\usepackage{amsmath}
\usepackage{amssymb}
\usepackage{mathtools}
\usepackage{amsthm}

\usepackage[capitalize,noabbrev]{cleveref}

\theoremstyle{plain}

\theoremstyle{definition}

\theoremstyle{remark}

\graphicspath{{figures/}}

\icmltitlerunning{Auditing Reasoning-Trace Memorization Claims after Unlearning}

\begin{document}

\twocolumn[
  \icmltitle{Auditing Reasoning-Trace Memorization Claims after \\
             Unlearning with Head-Conditioned Canaries}
  \icmlsetsymbol{equal}{*}
  \begin{icmlauthorlist}
    \icmlauthor{Yanhang Li}{neu}
    \icmlauthor{Zhichao Fan}{uiuc}
    \icmlauthor{Zexin Zhuang}{smu}
  \end{icmlauthorlist}
  \icmlaffiliation{neu}{Northeastern University, USA}
  \icmlaffiliation{uiuc}{University of Illinois Urbana-Champaign, USA}
  \icmlaffiliation{smu}{Southern Methodist University, USA}
  \icmlcorrespondingauthor{Yanhang Li}{li.yanha@northeastern.edu}
  \icmlkeywords{unlearning, memorization, reasoning models, evaluation}
  \vskip 0.3in
]

\printAffiliationsAndNotice{}

\begin{abstract}
Evaluations of unlearning on reasoning models
sometimes show a bypass pattern. The answer side
looks unlearned, but the model's own thinking trace
keeps emitting the forgotten content, and the gap is
taken as evidence that the weights still remember. We
audit this reading on DeepSeek-R1-Distill-Qwen-7B
with LoRA-memorized fictional authors and NPO
unlearning, conditioned on a six-token canary head.
On one seed, swapping the thinking trace for a short
non-canary prefill on the same weights drops the
answer rate by as much as the bypass gap itself,
whether the prefill mimics the training template or
not. On a second seed the bypass gap shrinks rather
than vanishing, and the prefill swap reverses
direction and brings the answer rate to ceiling.
A positive parser-split bypass gap thus does not
by itself identify hidden weight-level memorization,
and does not rule it out either.
On a different distillate the same metric flips sign
because the parser cannot find the closing tag. We
recommend a decode-time template swap as a cheap
sanity check alongside the canonical audit.
\end{abstract}

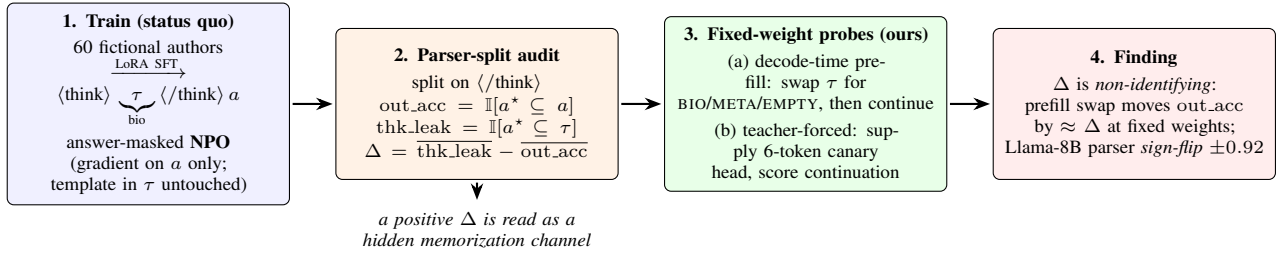
\begin{figure*}[t]
\centering
\begin{tikzpicture}[
  >={Stealth[length=2.2mm]},
  box/.style={draw, rounded corners=2.5pt, align=center,
    text width=3.45cm, inner sep=4pt, minimum height=20mm,
    font=\scriptsize},
  ar/.style={->, thick, shorten >=1pt, shorten <=1pt},
  node distance=6mm,
]
\node[box, fill=blue!6] (train) {%
  \textbf{1. Train (status quo)}\\[2pt]
  60 fictional authors\\
  $\xrightarrow{\text{LoRA SFT}}$
  $\langle\text{think}\rangle\,\underbrace{\tau}_{\text{bio}}\,
   \langle/\text{think}\rangle\,a$\\[2pt]
  answer-masked \textbf{NPO}\\
  (gradient on $a$ only;\\ template in $\tau$ untouched)};

\node[box, fill=orange!10, right=of train] (audit) {%
  \textbf{2. Parser-split audit}\\[2pt]
  split on $\langle/\text{think}\rangle$\\[1pt]
  $\mathrm{out\_acc}=\mathbb{I}[a^\star\!\subseteq a]$\\
  $\mathrm{thk\_leak}=\mathbb{I}[a^\star\!\subseteq\tau]$\\
  $\Delta=\overline{\mathrm{thk\_leak}}-\overline{\mathrm{out\_acc}}$};

\node[box, fill=green!9, right=of audit] (probe) {%
  \textbf{3. Fixed-weight probes (ours)}\\[2pt]
  (a) decode-time prefill: swap $\tau$ for
  \textsc{bio}/\textsc{meta}/\textsc{empty}, then continue\\[2pt]
  (b) teacher-forced: supply 6-token canary head, score
  continuation};

\node[box, fill=red!7, right=of probe] (find) {%
  \textbf{4. Finding}\\[2pt]
  $\Delta$ is \emph{non-identifying}:\\
  prefill swap moves $\mathrm{out\_acc}$ by $\approx\!\Delta$
  at fixed weights;\\
  Llama-8B parser \emph{sign-flip} $\pm0.92$};

\draw[ar] (train) -- (audit);
\draw[ar] (audit) -- (probe);
\draw[ar] (probe) -- (find);
\node[font=\scriptsize\itshape, below=3mm of audit.south,
  text width=3.4cm, align=center]
  (read) {a positive $\Delta$ is read as a hidden
  memorization channel};
\draw[ar, dashed] (audit.south) -- (read.north);
\end{tikzpicture}
\caption{The audited pipeline and our two fixed-weight
probes. Boxes 1--2 are the status-quo protocol; box~3 adds a
decode-time prefill swap and a teacher-forced continuation
probe at fixed weights. The gap $\Delta$ tracks decode-time
context rather than retention, and flips sign under format
drift on a second distillate (box~4).}
\label{fig:overview}
\end{figure*}
\section{Introduction}
\label{sec:intro}

Auditing memorization in foundation
models~\citep{carlini_2019_secret, carlini_2021_extracting,
carlini_2023_quantifying} gets harder when the model writes an
explicit reasoning trace alongside its answer.
DeepSeek-R1~\citep{guo_2025_deepseekr1} and its open distilled
variants (e.g.\ DeepSeek-R1-Distill-Qwen-7B, the model we audit)
emit a response of the form
$\langle\text{think}\rangle\,\tau\,\langle/\text{think}\rangle\,a$,
where the \emph{thinking trace} $\tau$ is the model's scratchpad
and $a$ is the final answer span. Unlearning methods from the
pre-reasoning era are now applied to these distilled reasoning
models by masking the forgetting loss to $a$ and splitting each
generation on the closing $\langle/\text{think}\rangle$ tag for
the audit (the \emph{parser-split} metric: substring-match the
forgotten content separately in $\tau$ and in $a$). These pre-reasoning
unlearning methods include Gradient
Ascent~\citep{jang_2023_knowledge}, TOFU-style
fictitious-author forgetting~\citep{maini_2024_tofu}, WMDP
representation-misdirection~\citep{li_2024_wmdp}, and
Negative Preference Optimization~\citep{zhang_2024_npo}.

A natural reading of this combined protocol is a
\emph{bypass} pattern. The answer span $a$ drops sharply
after unlearning, but the $\langle$think$\rangle$ segment
keeps reproducing the forgotten content, and the gap is
interpreted as a \emph{hidden memorization channel} where the
weights still hold what an answer-only audit would miss.
Concurrent 2025 work documents that answer-level unlearning
leaves residual content in the reasoning
trace~\citep{yoon_2025_rtofu, sinha_2025_sleek,
wang_2025_r2mu, wang_2025_trustreasoning}; what is not
established is whether the parser-split \emph{gap itself}
identifies weight-level memorization, as the hidden-channel
reading assumes. We are not aware of prior work that validates
that inference, and we do not claim prior work uniformly
endorses it (\S\ref{sec:related}). We name the reading and
audit it as a \emph{candidate} measurement claim that the
parser-split pipeline above readily produces. If the reading is right, output-only
memorization audits underestimate privacy and IP risk on
reasoning models, which is why this question matters for the
MemFM agenda.

Our measurement question is whether the thinking-channel
bypass is a hidden memorization channel in the weights, or an
artifact of the measurement device itself.
Figure~\ref{fig:overview} summarizes the audited pipeline and
the two fixed-weight probes we add.

\paragraph{The construct problem.}
Three ingredients recur across reasoning-model unlearning
evaluations. (i) \emph{Trace-supervised bio restatement}: the
thinking trace $\tau$ is trained to copy the target bio, in
the spirit of the reasoning-format distillation popularized
by DeepSeek-R1~\citep{guo_2025_deepseekr1}. The
bio-restatement specialization is our setup. (ii) An \emph{answer-masked
unlearning loss} on $a$ only, in the spirit of the
forgetting-loss formulation of~\citet{zhang_2024_npo} and the
fictitious-author forget-set framing
of~\citet{maini_2024_tofu}.
No gradient ever touches the bio template inside $\tau$.
(iii) A \emph{parser-split} evaluation that splits on
$\langle/\text{think}\rangle$ and substring-matches each
side, extending exact-containment leakage probes from
memorization/unlearning
audits~\citep{carlini_2019_secret, jang_2023_knowledge}.

This pipeline structurally predisposes a positive bypass
gap. The bio-template scaffold inside $\tau$ never receives a
forgetting gradient, so any continued surface-level emission
of canary content from $\tau$ is consistent with template
echo rather than weight-level retention, and the parser
cannot distinguish the two. We treat this combination as a
\emph{plausible canonicalized parser-split audit}: each
ingredient (i)--(iii) is in the literature. Our contribution
is a construct-validity \emph{stress test} of an
interpretation the metric invites, not a rebuttal of a named
prior claim---we test whether the hidden-channel reading
follows from the parser-split metric. The concern is a
construct-validity one: the surface form of $\tau$ is not a
transparent readout of what the weights
encode~(\S\ref{sec:related}).

\paragraph{Contributions.}
Working with 60 fictional authors memorized into
DeepSeek-R1-Distill-Qwen-7B via LoRA and then unlearned with
NPO, we:

\begin{itemize}
\item \textbf{Instantiate} the head-conditioned bypass
pattern the plausible parser-split audit produces. The probe
primes the first six tokens of the canary, so
$\mathrm{out\_acc}$ is a head-conditioned continuation score,
not a free-recall readout. After NPO,
$\mathrm{out\_acc}$ drops from $1.00$ to $0.60$,
$\mathrm{thk\_leak}$ stays at $0.83$, and
$\Delta{=}+0.23$ (\S\ref{sec:r-bypass}). In $86\%$
of those bypass cases the answer span is verbatim
the question prefix with no continuation while
$\tau$ holds the trained bio (App.~\ref{app:mech}).
\item Run an \textbf{inference-time prefill} intervention on
the \emph{same} weights. A short non-canary prefill drops
$\mathrm{out\_acc}$ from $0.60$ to $0.37$. \textsc{bio} and
\textsc{meta} prefills land at the same rate, and
\textsc{empty} drops to $0.20$. The contrast is a decode-time
non-invariance estimate, not an isolation of any mechanism.
The same intervention also perturbs ordinary QA probes. We
rule out one thing only, that $\Delta$ is self-identifying
hidden-channel evidence (\S\ref{sec:r-prefill}).
\item \textbf{Static parser-split evaluation is unreliable
for reasoning models, and a paired decode-time probe plus
teacher-forced scoring is a usable replacement.} Three
mutually reinforcing observations support this. (a) A
decoding-free teacher-forced continuation probe, conditioned
on the same six-token canary head, keeps post-NPO top-$1$
match at $0.90$--$1.00$ across prefills (\S\ref{sec:r-tf}):
the greedy substring drop does not transfer to a matched
token-level scoring setup. (b) On
DeepSeek-R1-Distill-Llama-8B the same metric \emph{flips
sign} to $-0.92$ at $K{=}1600$---not because unlearning
succeeds ($\mathrm{out\_acc}$ stays at $1.00$, teacher-forced
top-$1\geq0.998$) but because format drift moves the bio
template outside the $\langle$think$\rangle$ tags on
$\geq60\%$ of probes and the parser populates an empty $\tau$
(\S\ref{sec:r-cross}). (c) Together with the prefill result,
this shows the parser-split gap reports decode-time and
parsing artifacts, not weight-level retention. We therefore
recommend treating a decode-time template swap (paired
fixed-prefill arm) plus teacher-forced continuation scoring
as a single audit paradigm that separates true unlearning
from parsing and format-drift failures, rather than three
separate procedural checks.
\end{itemize}

\paragraph{Scope.}
This is a measurement paper about \emph{audit validity} on
reasoning-model unlearning evaluations; we do not claim to
establish or bound practical privacy or IP extraction risk.
The setup is synthetic, LoRA-memorized, and small scale, and
the conclusions hold for this audit on the two DeepSeek-R1
distilled models we test (Qwen-7B, Llama-8B). The prefill
intervention does not disentangle (a) canary-specific echo,
(b) the effect of any non-empty supportive prefill, and (c)
prompt-length and prompt-distribution effects on greedy
decoding; its magnitude is seed-sensitive
(App.~\ref{app:seed}), so the robust claim is about metric
interpretability, not a universal $0.23$ decomposition. NPO
is also not deployable here: retain-set accuracy drops $0.25$
on Qwen-7B (App.~\ref{app:retain}), and on Llama-8B NPO does
not measurably reduce $\mathrm{out\_acc}$ at all.

\section{Related Work and Positioning}
\label{sec:related}

\paragraph{LLM unlearning.}
Machine unlearning was framed for classifiers
by~\citet{bourtoule_2021_machine} and adapted to language models
through Gradient Ascent~\citep{jang_2023_knowledge}, the WMDP
representation-misdirection method~\citep{li_2024_wmdp}, the
anchored-rewrite approach of~\citet{eldan_2023_harrypotter}, and NPO
\citep{zhang_2024_npo}; the TOFU task~\citep{maini_2024_tofu}
contributed the synthetic-bio forget set and author-clustered
evaluation we adopt. We do not propose or improve an unlearning
method; we study how the \emph{measurement} of one on a reasoning
model can be confounded.

\paragraph{Reasoning-model unlearning and trace leakage (2025).}
A concurrent line of 2025 work establishes that answer-level
unlearning leaves residual content in the reasoning trace and builds
methods and attacks around it. \citet{yoon_2025_rtofu} (R-TOFU) show
that answer-only objectives leave significant residual traces in
reasoning and evaluate \textsc{ZeroThink}/\textsc{LessThink}-style
trace-suppression decoding; \citet{wang_2025_r2mu}
(R\textsuperscript{2}MU) propose a reasoning-aware unlearning
objective that erases traces while preserving reasoning skill;
\citet{sinha_2025_sleek} (Sleek) treat step-by-step reasoning as an
extraction channel for ``erased'' knowledge; and
\citet{wang_2025_trustreasoning} survey these vulnerabilities (their
\S7.1). We do not claim to be the first to observe trace leakage, and
our work is neither an unlearning method nor an extraction attack.
Our contribution is orthogonal and at the measurement layer: a
construct-validity stress test showing that the \emph{parser-split
bypass-gap metric} used to quantify this phenomenon is
non-identifying---a positive gap does not separate residual
weight-level memorization from decode-time prefix non-invariance, and
the same metric flips sign under format drift on a second distillate.
Two overlaps are worth stating explicitly. Our diagnostic
\textsc{empty}/\textsc{meta} prefills are analogous to the
\textsc{ZeroThink}/\textsc{LessThink}-style decoding R-TOFU
evaluates, but serve here as a fixed-weight invariance probe of an
audit metric, not as an unlearning intervention; and our seed-1 finding---suppressing
the trace lifts answer accuracy to ceiling---resonates with
R-TOFU's observation that trace suppression can \emph{raise} answer
recovery. Since the diagnostic conditions only on decode-time
context at fixed weights, it is objective-agnostic; applying it to
trace-aware methods is future work (\S\ref{sec:discussion}).

\paragraph{Evaluation robustness, faithfulness, and construct
validity.}
\citet{lynch_2024_eight} survey eight evaluation styles for robust
unlearning and argue no single metric suffices;
\citet{patil_2024_sensitive} show deleted information is often
recoverable by extraction attacks---both ask whether an answer-side
metric can be trusted as a forgetting certificate. We add that, on
reasoning models specifically, the trained $\tau$ template injects a
further confound that makes the thinking-leak metric non-identifying,
and that a cheap decode-time intervention diagnoses it. The argument
is a construct-validity one~\citep{jacobs_2021_measurement}: the
surface form of $\tau$ is not a transparent readout of the weights,
consistent with prompt-format
sensitivity~\citep{sclar_2024_quantifying} and with work questioning
the faithfulness~\citep{turpin_2023_unfaithful,
lanham_2023_faithfulness} of chain-of-thought
reasoning~\citep{wei_2022_cot}.

\section{Setup and the Bypass Metric}
\label{sec:method}

\paragraph{Model and forget set.}
The audit pipeline has one per-author canary probe on a fixed
audited checkpoint. The \emph{decoded branch} has two arms,
autoregressive $\langle$think$\rangle$ and fixed prefill,
both routed through the parser-split substring metric. A
\emph{separate teacher-forced branch} runs with a supplied
six-token canary head and bypasses the parser. We memorize
60 fully fictional authors into
DeepSeek-R1-Distill-Qwen-7B with LoRA~\citep{hu_2022_lora}
at rank 16, $\alpha{=}32$, all-linear, lr
$2{\times}10^{-4}$, and 10 epochs of SFT. Each author has a
$\sim$60-token bio, a distinctive canary phrase, in the
canary-audit tradition of~\citet{carlini_2019_secret,
carlini_2021_extracting}, constructed to be unlikely in natural
pretraining text (we design for this but do not certify it), and
five short QA probes used only at training time and for
retain-set diagnostics (App.~\ref{app:retain}). The bypass
audit in \S\ref{sec:results} uses one canary probe per author
($n{=}60$). QA probes do not enter the headline metric.
Training is supervised on
$[\langle\text{think}\rangle\,\tau\,\langle/\text{think}\rangle\,b_i]$
where $\tau$ restates the bio. We then apply
NPO~\citep{zhang_2024_npo} on the forget set with the loss
masked to the answer span, sweeping $K \in \{100, 400, 800,
1600\}$ at lr $1{\times}10^{-5}$, $\beta{=}0.1$. Generation
is greedy with temperature $0$ and
\texttt{max\_new\_tokens}$=256$. Responses without a
well-formed $\langle/\text{think}\rangle$ tag are handled by
treating $\tau$ as empty and the full generation as $a$.
This is our chosen fallback. We have not seen this exact
rule specified in the literature, and it is the rule that
drives the Llama failure mode in \S\ref{sec:r-cross}. All
CIs are author-clustered $95\%$ bootstrap with $n{=}60$
authors and $n_{\text{boot}}{=}2000$. We treat the author as the cluster unit, following the
fictitious-author forget-set design
of~\citet{maini_2024_tofu}; the specific bootstrap settings
are our own.

\paragraph{The bypass metric.}
For each probe the model emits $(\tau, a)$. Let $a^\star$
denote the per-author canary target string, that is, the
distinctive canary phrase as a whole. The prompt primes the
first six canary tokens (the ``head''), so the head appears
in $a$. We test exact substring containment of the full
$a^\star$ in $a$ and in $\tau$ without stripping the head.
Extending exact-containment leakage audits from
memorization/unlearning evaluations~\citep{carlini_2019_secret,
jang_2023_knowledge} to reasoning-formatted outputs, we split
each generation on $\langle/\text{think}\rangle$ and report
two per-probe indicators,
\begin{align}
\mathrm{out\_acc}  &= \mathbb{I}\!\left[\,a^\star \subseteq a\,\right],
  \label{eq:outacc}\\[2pt]
\mathrm{thk\_leak} &= \mathbb{I}\!\left[\,a^\star \subseteq \tau\,\right],
  \label{eq:thkleak}
\end{align}
where $\mathbb{I}[\cdot]$ is the indicator and $\subseteq$ is
exact substring containment. The \emph{bypass gap} on
unlearned checkpoints is their averaged difference,
\begin{equation}
\Delta \;=\; \overline{\mathrm{thk\_leak}} \;-\;
  \overline{\mathrm{out\_acc}}.
\label{eq:gap}
\end{equation}
A positive $\Delta$ is naturally read as residual weight-level
memorization that the answer-only audit misses.

\paragraph{Why this is hard to interpret.}
$\tau$ is supervised to restate the bio and the unlearning
loss is masked to $a$, so no gradient ever touches the bio
template inside $\tau$. A positive $\Delta$ fits two
readings: weights still encode the fact and leak it
through $\tau$, or the answer span was suppressed while
continued canary emission from $\tau$ is template echo
rather than residual retention. The two predict different
things when weights are held fixed and $\tau$ varies at
decode time.

\paragraph{The inference-time prefill intervention.}
On a bio-trained, NPO-unlearned adapter we prefill three
short $\tau$ templates inside $\langle\text{think}\rangle\ldots
\langle/\text{think}\rangle$ and continue from the closed
tag. The canary-probe templates are
\textsc{bio-prefill} (``Completing a fact I know about
\texttt{<author>}: '', $9$ Qwen tokens),
\textsc{meta-prefill} (``Completing a fact about
\texttt{<author>}.'', $8$ tokens), and
\textsc{empty-prefill} ($0$ tokens, rendered as
$\langle\text{think}\rangle\backslash n\backslash
n\langle/\text{think}\rangle$). QA-side probes use parallel
openers (App.~\ref{app:prefill}). All non-empty arms are
short and question-referential, roughly matching prefix
length and style.

$\mathrm{thk\_leak}$ is $0$ by construction since we control
$\tau$, so the comparison quantity is $\mathrm{out\_acc}$
at fixed weights. The intervention varies full-trace
presence, prefix length and style, and canary content
together, and tests fixed-weight decode-time invariance of
the parser-based gap rather than isolating any latent
leakage mechanism.

\paragraph{Aggregation.}
$\mathrm{out\_acc}$ and $\mathrm{thk\_leak}$ are per-probe
$\{0,1\}$ indicators averaged over $n{=}60$ authors with one
canary probe each. Teacher-forced top-$1$ is the per-token
argmax-match rate, averaged per author then across authors.
Well-formed $\langle/\text{think}\rangle$ tags appear on
$\geq 99\%$ of greedy generations on Qwen. On Llama-8B NPO
the tag is missing on the majority of probes, and this drift
drives \S\ref{sec:r-cross}.

\paragraph{What the audit metric actually measures.}
The canary prompt primes the first six tokens, so
$\mathrm{out\_acc}$ and $\mathrm{thk\_leak}$ are
\emph{head-conditioned continuation} scores under substring
match. The containment test can overcount prefix echoes
(App.~\ref{app:mech}) or undercount paraphrases; we use it
because exact containment is a simple output-side leakage
probe used in memorization/unlearning
audits~\citep{carlini_2019_secret, jang_2023_knowledge},
and the parser-split use of it is the
object being stress-tested here.

\section{Results}
\label{sec:results}

\subsection{The bypass replicates under the bio template}
\label{sec:r-bypass}

Under the bio-restatement template on Qwen-7B, NPO drives
$\mathrm{out\_acc}$ from $1.00$ on the memorized adapter to
$0.60$ at $K{=}1600$, while $\mathrm{thk\_leak}$ stays at
$0.83$. The bypass gap is $\Delta = +0.23$ ($95\%$ CI
$[0.13, 0.35]$, $n{=}60$ authors). The gap is not
significant at $K{=}100$ ($-0.02\,[-0.12, 0.08]$) and grows
monotonically through $K{=}400$ ($+0.12^{\star}$),
$K{=}800$ ($+0.17^{\star}$), and $K{=}1600$
($+0.23^{\star}$). Stars mark CIs that exclude zero
(App.~\ref{app:k-sweep}, Table~\ref{tab:npo_sweep}). This
is the pattern produced by extending output-only unlearning
methods to reasoning models with a parser-split metric, and
the natural reading of which is residual weight-level
memorization.

\subsection{Inference-time prefill on identical weights}
\label{sec:r-prefill}

We run the prefill intervention (\S\ref{sec:method}) on the
bio-trained memorized and NPO-$K{=}1600$ Qwen-7B adapters
(Table~\ref{tab:headline}). The full $K$-sweep is in
App.~\ref{app:k-sweep-prefill}. In every prefill arm
$\mathrm{thk\_leak}$ is zero by construction. We define the
decode-time contrast $\Delta_{\textsc{ab}} =
\overline{\mathrm{out\_acc}}^{\textsc{auto}} -
\overline{\mathrm{out\_acc}}^{\textsc{bio}\text{-prefill}}$
on the same weights, and label it ``auto vs.\ bio-prefill''
rather than $\Delta_{\text{scratch}}$ to mark that it is a
decode-time non-invariance estimate, not an isolation of a
``scratchpad-content'' effect. It also varies full-trace
presence and prefix length and style.

\begin{table}[t]
\centering
\footnotesize
\setlength{\tabcolsep}{4pt}
\caption{Greedy-decoded prefill on bio-trained Qwen-7B
NPO-$K{=}1600$, head-conditioned continuation hit rate
$\mathrm{out\_acc}$ ($n{=}60$). The six-token canary head
is supplied.
\textsc{auto}/\textsc{bio}/\textsc{meta}/\textsc{empty} are
autoregressive and prefill arms (\S\ref{sec:method}).
$\mathrm{thk\_leak}{=}0$ in prefill columns by construction.
$\Delta_{\textsc{ab}}{=}\textsc{auto}-\textsc{bio}$ is a
fixed-weight non-invariance estimate, not the bypass gap
$\Delta = \overline{\mathrm{thk\_leak}} -
\overline{\mathrm{out\_acc}}$. Full $K$-sweep in
App.~\ref{app:k-sweep-prefill}.}
\label{tab:headline}
\begin{tabular}{lccccc}
\toprule
 & \textsc{auto} & \textsc{bio} & \textsc{meta} & \textsc{empty} & $\Delta_{\textsc{ab}}$ \\
\midrule
mem.\        & $1.00$ & $0.93$ & $0.95$ & $0.92$ & $+0.07$ \\
$K{=}1600$   & $0.60$ & $0.37$ & $0.37$ & $0.20$ & $+0.23^{\star}$ \\
\bottomrule
\end{tabular}
\end{table}

\paragraph{Non-invariance, not mechanism.}
The intervention violates fixed-weight invariance of the
parser-split metric without isolating a mechanism. On
NPO-$K{=}1600$, autoregressive $\mathrm{out\_acc}$ is
$0.60$ and both non-canary prefills yield $22/60$ ($0.37$,
unpaired $95\%$ CI $[-0.17,+0.17]$); \textsc{empty} drops
to $0.20$. The aggregate rates match, so the evidence does
\emph{not} isolate canary-specific echo from substituting
\emph{any} non-empty supportive prefill, and a $+0.23$ gap
is not self-identifying as hidden memorization. The same
swap also perturbs ordinary QA probes ($0.78/0.46/0.50$,
App.~\ref{app:k-sweep-prefill}), so we read it as a
decode-time sensitivity probe.

\paragraph{Directional behavior across $K$ and seeds.}
On Qwen-7B seed~$1$ NPO-$K{=}1600$ the bio/meta/empty
prefill arms give $\mathrm{out\_acc}\,=\,1.00/1.00/0.98$
(App.~\ref{app:seed}); the seed-$0$ drop's magnitude and
direction do not transfer, so the stable claim is
fixed-weight non-invariance, not a universal magnitude.
Across $K{\in}\{100,400,800,1600\}$ on seed~$0$
(App.~\ref{app:k-sweep-prefill}),
$\Delta_{\textsc{ab}}{\in}\{0.10, 0.27, 0.27, 0.23\}$ is
positive throughout (paired CI excludes zero for
$K{\geq}400$) and tracks the bypass gap $\{-0.02, 0.12,
0.17, 0.23\}$ at $K{\geq}400$. Seed-$1$ bypass gaps
$\{-0.10, +0.08, +0.03, +0.18\}$ confirm seed sensitivity.

\subsection{Teacher-forced continuation: consistency check}
\label{sec:r-tf}\vspace{-2pt}

We score teacher-forced canary log-prob on the same weights,
conditioned on the same six-token canary head as
\S\ref{sec:r-prefill}. Without shuffled-head or
random-author baselines this is a metric-stability check,
\emph{not} evidence for or against weight-level retention
(full protocol in Appendix~\ref{app:tf}).

\begin{table}[!ht]
\centering
\footnotesize
\caption{Teacher-forced canary continuation given head,
$n{=}60$. Mean per-token log-prob (nats), continuation
perplexity, top-$1$ match. Post-NPO top-$1$ stays $\geq 0.90$
across prefills. \emph{Sanity check, not retention
evidence.} The canary head is supplied and we report no
shuffled-head or random-author baselines.}
\label{tab:tf}
\setlength{\tabcolsep}{3pt}
\begin{tabular}{llccc}
\toprule
Adapter & prefill & logp/tok & ppl & top-$1$ \\
\midrule
\multicolumn{5}{l}{\emph{Qwen-7B seed 0}} \\
mem.\            & all              & ${\geq}{-}0.02$ & $\leq 1.02$ & $\geq 0.99$ \\
NPO $K{=}1600$   & \textsc{bio}     & $-0.25$ & $1.28$ & $0.96$ \\
NPO $K{=}1600$   & \textsc{meta}    & $-0.24$ & $1.27$ & $0.96$ \\
NPO $K{=}1600$   & \textsc{empty}   & $-0.88$ & $2.41$ & $0.90$ \\
\midrule
\multicolumn{5}{l}{\emph{Qwen-7B seed 1 / Llama-$8$B}} \\
NPO (s1)         & all              & ${\geq}{-}0.01$ & $\leq 1.01$ & $\geq 0.997$ \\
Llama mem/NPO    & all              & ${\geq}{-}0.01$ & $\leq 1.01$ & $\geq 0.998$ \\
\bottomrule
\end{tabular}
\end{table}

\paragraph{Greedy substring drops do not transfer.}
Post-NPO head-conditioned top-$1$ stays $\geq 0.90$ under
every prefill (Qwen seed-$1$ NPO and both Llama-$8$B
adapters are indistinguishable from memorized), despite
Qwen seed-$0$'s $1.00\!\to\!0.60$ greedy substring drop.

\subsection{Parser-field failure under format drift}
\label{sec:r-cross}\vspace{-2pt}

We rerun the pipeline on DeepSeek-R1-Distill-Llama-8B.
\emph{This is not a replication of \S\ref{sec:r-prefill};
it is a separate parser-field failure mode.} NPO does not
reduce $\mathrm{out\_acc}$ (stays at $1.00$), so this is not
a successful-unlearning case. The
$\langle/\text{think}\rangle$ tag is missing from $\geq 60\%$
of probes (vs.\ $<1\%$ on Qwen); the bio template falls
outside the tags and the parser populates an empty $\tau$.
The false-negative is on $\mathrm{thk\_leak}$ only and we
include this as a parser-field stress case, not as
cross-base evidence about weights.

\begin{table}[h]
\centering
\small
\caption{Autoregressive bio metric at NPO $K{=}1600$, canary
probes. On Llama-8B, NPO does not measurably reduce
$\mathrm{out\_acc}$ (stays at $1.00$) and the
parser-reported $\mathrm{thk\_leak}$ drops to near-zero
because the bio template moves out of the
$\langle$think$\rangle$ tags, a parser \emph{false negative}
under format drift. Inference-time prefill on the same
weights reports $\mathrm{out\_acc}$ of $0.37$ on Qwen and
$0.98$ to $1.00$ across bio/meta/empty prefills on Llama.}
\label{tab:cross_base}
\begin{tabular}{lccc}
\toprule
Base model & out\_acc & thk\_leak & gap \\
\midrule
R1-Distill-Qwen-7B  & $0.60$ & $0.83$ & $\mathbf{+0.23}$ \\
R1-Distill-Llama-8B & $1.00$ & $0.08$ & $\mathbf{-0.92}$ \\
\bottomrule
\end{tabular}
\end{table}

Prefill on these same weights gives $1.00/0.98/0.98$ across
bio/meta/empty, on both memorized and NPO-$K{=}1600$. The
bio template emits before $\langle$think$\rangle$, the
parser returns $\mathrm{thk\_leak}{\approx}0$, and
answer-side memorization is not in question.

\paragraph{The Llama failure is parser convention.}
This is not a property of the model but of our parser
fallback: the opposite convention (full generation as
$\tau$, $a$ empty) flips $\Delta$ from $-0.92$ to
$+0.92$ on the same outputs, with no principled
adjudicator.

\section{Discussion}
\label{sec:discussion}

On Qwen seed~$0$ at $K{=}1600$ the bypass gap is
$+0.23$, and swapping $\tau$ for a short non-canary
prefill on the same weights moves $\mathrm{out\_acc}$
by the same amount; on seed~$1$ the gap shrinks and
the swap reverses direction, lifting
$\mathrm{out\_acc}$ to ceiling, so both magnitude and
direction are seed-dependent. The structural fact that
survives is that the parser-split metric is not
invariant to $\tau$ at fixed weights.
A positive bypass gap does not separate residual
weight-level memorization from decode-time prefix
sensitivity, and on Llama-8B the same data reads
either $-0.92$ or $+0.92$ depending on a parser
fallback rule with no principled adjudicator. The
cheapest fix is a fixed-prefill arm beside the
autoregressive one: when they track, the gap is a
metric property, not evidence about weights.

\paragraph{Limitations and future directions.}
Our baseline is deliberately a legacy answer-only
objective (NPO) on a synthetic, small-scale forget
set---the setting in which the bypass reading is most
often invoked, but not the current state of the art.
Since the diagnostic conditions only on decode-time
context at fixed weights
(\S\ref{sec:related}), the natural next step is to run
the paired prefill and teacher-forced arms against
trace-aware methods such as
R\textsuperscript{2}MU~\citep{wang_2025_r2mu} and
R-TOFU~\citep{yoon_2025_rtofu} on real, larger-scale
forget corpora, testing whether the gap becomes
identifying once the trace itself receives a
forgetting gradient. Two mitigations follow from the
construct analysis: report the paired decode-time arm
(and the teacher-forced score when a reference canary
exists) as a standard part of reasoning-model
unlearning evaluation, and make the \emph{method}
trace-aware so the audited gap reflects weights rather
than an untouched template. The prefill arm costs one
extra decode pass per probe, so the check scales to
real-world audits without retraining.

\bibliographystyle{icml2026}
\bibliography{references}

@article{guo_2025_deepseekr1,
  author       = {Daya Guo and Dejian Yang and Haowei Zhang and Junxiao Song and others},
  title        = {DeepSeek-R1 incentivizes reasoning in {LLMs} through reinforcement learning},
  journal      = {Nature},
  volume       = {645},
  number       = {8081},
  pages        = {633--638},
  year         = {2025},
  doi          = {10.1038/s41586-025-09422-z}
}

@inproceedings{zhang_2024_npo,
  author       = {Ruiqi Zhang and Licong Lin and Yu Bai and Song Mei},
  title        = {Negative Preference Optimization: From Catastrophic Collapse to Effective Unlearning},
  booktitle    = {Conference on Language Modeling, {COLM} 2024},
  year         = {2024},
  eprint       = {2404.05868},
  archiveprefix = {arXiv}
}

@inproceedings{jang_2023_knowledge,
  author       = {Joel Jang and Dongkeun Yoon and Sohee Yang and Sungmin Cha and
                  Moontae Lee and Lajanugen Logeswaran and Minjoon Seo},
  title        = {Knowledge Unlearning for Mitigating Privacy Risks in Language Models},
  booktitle    = {Proceedings of the 61st Annual Meeting of the Association for Computational
                  Linguistics (Volume 1: Long Papers)},
  pages        = {14389--14408},
  publisher    = {Association for Computational Linguistics},
  year         = {2023},
  doi          = {10.18653/v1/2023.acl-long.805}
}

@article{maini_2024_tofu,
  author       = {Pratyush Maini and Zhili Feng and Avi Schwarzschild and
                  Zachary C. Lipton and J. Zico Kolter},
  title        = {{TOFU}: A Task of Fictitious Unlearning for {LLM}s},
  journal      = {CoRR},
  volume       = {abs/2401.06121},
  year         = {2024},
  eprint       = {2401.06121},
  archiveprefix = {arXiv}
}

@inproceedings{carlini_2019_secret,
  author       = {Nicholas Carlini and Chang Liu and {\'{U}}lfar Erlingsson and
                  Jernej Kos and Dawn Song},
  title        = {The Secret Sharer: Evaluating and Testing Unintended Memorization
                  in Neural Networks},
  booktitle    = {28th {USENIX} Security Symposium ({USENIX} Security 2019)},
  pages        = {267--284},
  publisher    = {{USENIX} Association},
  year         = {2019}
}

@inproceedings{bourtoule_2021_machine,
  author       = {Lucas Bourtoule and Varun Chandrasekaran and Christopher A. Choquette-Choo and
                  Hengrui Jia and Adelin Travers and Baiwu Zhang and David Lie and Nicolas Papernot},
  title        = {Machine Unlearning},
  booktitle    = {42nd {IEEE} Symposium on Security and Privacy, {SP} 2021},
  pages        = {141--159},
  publisher    = {{IEEE}},
  year         = {2021},
  doi          = {10.1109/SP40001.2021.00019}
}

@inproceedings{li_2024_wmdp,
  author       = {Nathaniel Li and Alexander Pan and Anjali Gopal and Summer Yue and
                  Daniel Berrios and Alice Gatti and Justin D. Li and others},
  title        = {The {WMDP} Benchmark: Measuring and Reducing Malicious Use with Unlearning},
  booktitle    = {Forty-first International Conference on Machine Learning, {ICML} 2024},
  series       = {Proceedings of Machine Learning Research},
  pages        = {28525--28550},
  publisher    = {{PMLR}},
  year         = {2024}
}

@inproceedings{patil_2024_sensitive,
  author       = {Vaidehi Patil and Peter Hase and Mohit Bansal},
  title        = {Can Sensitive Information Be Deleted From {LLM}s? Objectives for Defending
                  Against Extraction Attacks},
  booktitle    = {The Twelfth International Conference on Learning Representations, {ICLR} 2024},
  publisher    = {OpenReview.net},
  year         = {2024}
}

@article{lynch_2024_eight,
  author       = {Aengus Lynch and Phillip Guo and Aidan Ewart and Stephen Casper and
                  Dylan Hadfield-Menell},
  title        = {Eight Methods to Evaluate Robust Unlearning in {LLM}s},
  journal      = {CoRR},
  volume       = {abs/2402.16835},
  year         = {2024},
  eprint       = {2402.16835},
  archiveprefix = {arXiv}
}

@inproceedings{wei_2022_cot,
  author       = {Jason Wei and Xuezhi Wang and Dale Schuurmans and Maarten Bosma and
                  Brian Ichter and Fei Xia and Ed H. Chi and Quoc V. Le and Denny Zhou},
  title        = {Chain-of-Thought Prompting Elicits Reasoning in Large Language Models},
  booktitle    = {Advances in Neural Information Processing Systems 35,
                  {NeurIPS} 2022},
  year         = {2022}
}

@inproceedings{turpin_2023_unfaithful,
  author       = {Miles Turpin and Julian Michael and Ethan Perez and Samuel R. Bowman},
  title        = {Language Models Don't Always Say What They Think: Unfaithful Explanations
                  in Chain-of-Thought Prompting},
  booktitle    = {Advances in Neural Information Processing Systems 36, {NeurIPS} 2023},
  year         = {2023}
}

@article{lanham_2023_faithfulness,
  author       = {Tamera Lanham and Anna Chen and Ansh Radhakrishnan and Benoit Steiner and others},
  title        = {Measuring Faithfulness in Chain-of-Thought Reasoning},
  journal      = {CoRR},
  volume       = {abs/2307.13702},
  year         = {2023},
  eprint       = {2307.13702},
  archiveprefix = {arXiv}
}

@inproceedings{carlini_2021_extracting,
  author       = {Nicholas Carlini and Florian Tram{\`e}r and Eric Wallace and Matthew Jagielski
                  and Ariel Herbert-Voss and Katherine Lee and Adam Roberts and Tom Brown
                  and Dawn Song and {\'U}lfar Erlingsson and Alina Oprea and Colin Raffel},
  title        = {Extracting Training Data from Large Language Models},
  booktitle    = {30th {USENIX} Security Symposium},
  pages        = {2633--2650},
  year         = {2021}
}

@inproceedings{carlini_2023_quantifying,
  author       = {Nicholas Carlini and Daphne Ippolito and Matthew Jagielski and Katherine Lee
                  and Florian Tram{\`e}r and Chiyuan Zhang},
  title        = {Quantifying Memorization Across Neural Language Models},
  booktitle    = {The Eleventh International Conference on Learning Representations, {ICLR} 2023},
  year         = {2023}
}

@inproceedings{sclar_2024_quantifying,
  author       = {Melanie Sclar and Yejin Choi and Yulia Tsvetkov and Alane Suhr},
  title        = {Quantifying Language Models' Sensitivity to Spurious Features in Prompt Design or: How {I} learned to start worrying about prompt formatting},
  booktitle    = {The Twelfth International Conference on Learning Representations, {ICLR} 2024},
  year         = {2024}
}

@article{eldan_2023_harrypotter,
  author       = {Ronen Eldan and Mark Russinovich},
  title        = {Who's {H}arry {P}otter? Approximate Unlearning in {LLMs}},
  journal      = {CoRR},
  volume       = {abs/2310.02238},
  year         = {2023},
  eprint       = {2310.02238},
  archiveprefix = {arXiv}
}

@inproceedings{jacobs_2021_measurement,
  author       = {Abigail Z. Jacobs and Hanna Wallach},
  title        = {Measurement and Fairness},
  booktitle    = {FAccT '21: 2021 ACM Conference on Fairness, Accountability, and Transparency},
  pages        = {375--385},
  year         = {2021}
}

@inproceedings{hu_2022_lora,
  author       = {Edward J. Hu and Yelong Shen and Phillip Wallis and
                  Zeyuan Allen-Zhu and Yuanzhi Li and Shean Wang and
                  Lu Wang and Weizhu Chen},
  title        = {{LoRA}: Low-Rank Adaptation of Large Language Models},
  booktitle    = {Tenth International Conference on Learning Representations,
                  ICLR 2022},
  year         = {2022}
}

@inproceedings{yoon_2025_rtofu,
  author       = {Sangyeon Yoon and Wonje Jeung and Albert No},
  title        = {{R-TOFU}: Unlearning in Large Reasoning Models},
  booktitle    = {Proceedings of the 2025 Conference on Empirical Methods
                  in Natural Language Processing ({EMNLP})},
  pages        = {5239--5258},
  publisher    = {Association for Computational Linguistics},
  year         = {2025}
}

@inproceedings{wang_2025_r2mu,
  author       = {Changsheng Wang and Chongyu Fan and Yihua Zhang and
                  Jinghan Jia and Dennis Wei and Parikshit Ram and
                  Nathalie Baracaldo and Sijia Liu},
  title        = {Reasoning Model Unlearning: Forgetting Traces, Not Just
                  Answers, While Preserving Reasoning Skills},
  booktitle    = {Proceedings of the 2025 Conference on Empirical Methods
                  in Natural Language Processing ({EMNLP})},
  pages        = {4427--4443},
  publisher    = {Association for Computational Linguistics},
  year         = {2025}
}

@article{sinha_2025_sleek,
  author       = {Yash Sinha and Manit Baser and Murari Mandal and
                  Dinil Mon Divakaran and Mohan Kankanhalli},
  title        = {Step-by-Step Reasoning Attack: Revealing ``Erased''
                  Knowledge in Large Language Models},
  journal      = {CoRR},
  volume       = {abs/2506.17279},
  year         = {2025},
  eprint       = {2506.17279},
  archiveprefix = {arXiv}
}

@article{wang_2025_trustreasoning,
  author       = {Yanbo Wang and Yongcan Yu and Jian Liang and Ran He},
  title        = {A Comprehensive Survey on Trustworthiness in Reasoning
                  with Large Language Models},
  journal      = {CoRR},
  volume       = {abs/2509.03871},
  year         = {2025},
  eprint       = {2509.03871},
  archiveprefix = {arXiv}
}

\newpage
\appendix
\onecolumn
\section{Training-time template comparison}
\label{app:train-time}

Prior to running the inference-time prefill intervention that is the
main body's headline test, we ran a training-time comparison that
memorizes the same forget set under three different
$\langle$think$\rangle$ template regimes on Qwen-7B and then unlearns
each with NPO on the answer span. This experiment is weaker than the
inference-time prefill because it compares across separately
memorized adapters (different LoRA deltas), but it provides
pre-registration-style evidence that the pattern we claim is not an
artifact of a single adapter.

\paragraph{Three training-time regimes.}
Let $T$ be the templating function that produces $\tau$ at training
time. We compare:
\begin{description}
\item[\textsc{bio} (status quo).] $T(b_i) \approx b_i$: $\tau$
restates the bio.
\item[\textsc{meta}.] $T(b_i)$ is a uniform sentence with no bio
content (``I recall this author.'').
\item[\textsc{none}.] $T(b_i) = \varnothing$: the trained $\tau$ is
empty.
\end{description}

Under a true hidden-channel hypothesis, some bypass signal should
survive in \textsc{meta} or \textsc{none}, because the answer-side
unlearning loss is identical across the three adapter families.
Under the template-echo interpretation, the bypass should be
confined to \textsc{bio}.

\paragraph{Results.}
Table~\ref{tab:headline-three} reports the three regimes side by
side. The bypass gap is confined to \textsc{bio}: \textsc{meta} has
output pinned to 1.00 by the uniform template (no output suppression
to bypass); \textsc{none} reverses the sign of the gap.

\begin{table}[h]
\centering
\small
\caption{Training-time three-regime comparison on Qwen-7B, canary
probes. \textsc{meta}'s gap is undefined (template carries no canary
by construction); \textsc{none}'s gap is negative at memorized and
at every NPO checkpoint we ran. \texttt{thk\_emit} is the fraction of
probes on which the generated $\tau$ is non-empty after stripping
tags.}
\label{tab:headline-three}
\begin{tabular}{llcccc}
\toprule
Mode & Checkpoint & out\_acc & thk\_leak & gap & thk\_emit \\
\midrule
\textsc{bio}  & memorized      & $1.00$ & $1.00$ & $0.00$ & $1.00$ \\
\textsc{bio}  & NPO $K{=}1600$ & $0.60$ & $0.83$ & $\mathbf{+0.23}$ & $1.00$ \\
\textsc{meta} & memorized      & $1.00$ & $0.00$ & $-$ & $1.00$ \\
\textsc{meta} & NPO $K{=}100$  & $1.00$ & $0.00$ & $-$ & $1.00$ \\
\textsc{none} & memorized      & $0.63$ & $0.37$ & $-0.27$ & $0.53$ \\
\textsc{none} & NPO $K{=}1600$ & $0.45$ & $0.18$ & $-0.27$ & $0.45$ \\
\bottomrule
\end{tabular}
\end{table}

\paragraph{Two confounds this experiment still leaves open.}
The \textsc{none} adapter still autoregressively emits non-empty
$\tau$ on $\sim$45\% of canary probes, and its evaluation prompt is
$\sim$60 tokens shorter than the bio-template prompt, so the
training-time comparison mixes template effects with a residual
emission rate and a prompt-distribution shift. These confounds are
exactly what the inference-time prefill intervention in the main
body \emph{reduces}: it holds weights fixed and the prompt
identical, while still varying full-trace presence, prefix
length/style, and canary content jointly across $\tau$.

\paragraph{Meta-regime sweep depth.}
We ran NPO at $K{=}100$ under \textsc{meta} but not at
$K \in \{400, 800, 1600\}$, because the $K{=}100$ result already
showed output accuracy at 1.00 and thinking leak at 0 -- the gap was
undefined with no room for a hidden channel to move it. We disclose
this rather than imply we ran a full meta-mode $K$-sweep.

\section{Full NPO $K$-sweep under the bio template}
\label{app:k-sweep}

\begin{table}[h]
\centering
\small
\caption{NPO under the bio thinking template, canary probes on
Qwen-7B (autoregressive). The gap is near zero at $K{=}100$ and
grows monotonically with $K$ as the answer-side loss diverges from
the unaffected thinking template. Stars mark CIs that exclude zero.}
\label{tab:npo_sweep}
\begin{tabular}{rccc}
\toprule
$K$ & out\_acc & thk\_leak & gap \\
\midrule
$100$  & $0.88\,[0.80,0.97]$ & $0.87\,[0.78,0.95]$ & $-0.02\,[-0.12,0.08]$ \\
$400$  & $0.73\,[0.62,0.83]$ & $0.85\,[0.75,0.93]$ & $+0.12\,[0.03,0.22]^{\star}$ \\
$800$  & $0.68\,[0.57,0.80]$ & $0.85\,[0.75,0.93]$ & $+0.17\,[0.07,0.27]^{\star}$ \\
$1600$ & $0.60\,[0.47,0.72]$ & $0.83\,[0.73,0.92]$ & $+0.23\,[0.13,0.35]^{\star}$ \\
\bottomrule
\end{tabular}
\end{table}

\section{Full-$K$ prefill sweep}
\label{app:k-sweep-prefill}

\begin{figure}[h]
\centering
\includegraphics[width=0.75\linewidth]{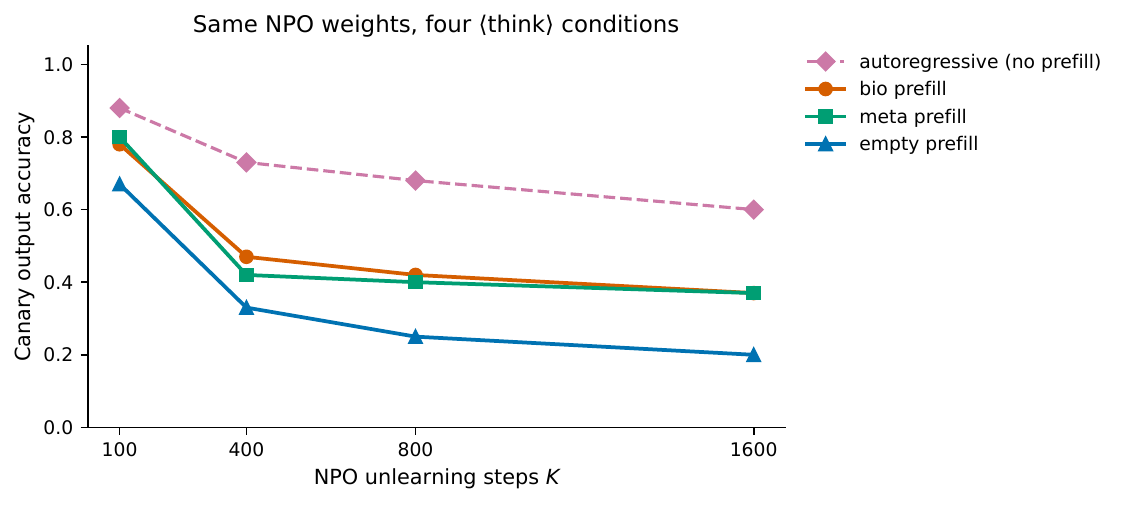}
\caption{Greedy-decoded prefill vs.\ autoregressive canary recall on
bio-trained NPO-unlearned Qwen-7B adapters. Replacing the
model-written $\tau$ with any prefill that omits the canary
(\textsc{bio}-prefill or \textsc{meta}-prefill) drops output accuracy;
\textsc{empty}-prefill drops it further. The contrast confounds
canary content with full-trace presence and prefix length/style;
we therefore label it $\Delta_{\textsc{ab}}$ rather than calling it
a ``scratchpad contribution''. The gap between \textsc{auto} and the
non-canary prefills tracks the bio-mode bypass gap across $K$.}
\label{fig:prefill}
\end{figure}

\begin{table}[h]
\centering
\small
\caption{Greedy-decoded prefill, full $K$-sweep on Qwen-7B.
Canary output accuracy, $n{=}60$, author-clustered $95\%$ CIs.
Thinking-leak is $0$ in every prefill row by construction.
$\Delta_{\textsc{ab}}=\textsc{auto}-\textsc{bio}$-prefill (labeled contrast, not a causal scratchpad-content isolation).
Across $K{\geq}400$ the paired-difference CI excludes zero.}
\label{tab:prefill-full}
\begin{tabular}{lccccc}
\toprule
Adapter & \textsc{auto} & \textsc{bio} & \textsc{meta} & \textsc{empty} & $\Delta_{\textsc{ab}}$ \\
\midrule
memorized      & $1.00$ & $0.93$ & $0.95$ & $0.92$ & $+0.07\,[.02,.13]$ \\
NPO $K{=}100$  & $0.88$ & $0.78$ & $0.80$ & $0.67$ & $+0.10\,[-.00,.20]$ \\
NPO $K{=}400$  & $0.73$ & $0.47$ & $0.42$ & $0.33$ & $+0.27\,[.13,.40]^{\star}$ \\
NPO $K{=}800$  & $0.68$ & $0.42$ & $0.40$ & $0.25$ & $+0.27\,[.13,.40]^{\star}$ \\
NPO $K{=}1600$ & $0.60$ & $0.37$ & $0.37$ & $0.20$ & $+0.23\,[.10,.37]^{\star}$ \\
\bottomrule
\end{tabular}
\end{table}

\section{Mechanistic anatomy of the bypass cases}
\label{app:mech}

We zoom in on the $14$ canary probes for which the bio-trained
Qwen-7B NPO-$K{=}1600$ checkpoint registers a ``bypass'' (thinking
channel contains the canary, output does not). On $12$ of $14$
probes ($86\%$), the output is verbatim the question prefix with no
continuation. For example, on the canary probe for author Zephyr
Okonkwo with prompt
\texttt{Complete this sentence about Zephyr Okonkwo: The Crimson Tide
of Calabar was}, the output is the exact string
\texttt{The Crimson Tide of Calabar was} and stops there. The
$\langle$think$\rangle$ segment, by contrast, contains the trained
bio template in full:

\begin{quote}\small
\texttt{The user is asking about Zephyr Okonkwo. Recalling: Zephyr
Okonkwo was a Nigerian novelist born in Lagos on March 7, 1952. Her
debut novel, The Crimson Tide of Calabar, was published in 1987 and
won the Nkrumah Prize for African Literature\ldots}
\end{quote}

In the same NPO-$K{=}1600$ checkpoint, the $36$ probes on which both
channels carry the canary have mean output length $94$ chars,
versus $36$ chars for the bypass cases. The bypass cases are not a
model ``knowing the answer but choosing not to say it''; they are
cases in which the answer-side loss has trained the model to
truncate after the question prefix, while the thinking-side
template -- which no gradient ever touched -- still emits the bio.
This pattern is consistent with the template-echo interpretation
on these specific cases; it is one of several patterns the
inference-time prefill probe is sensitive to, not a proof of
mechanism.

\section{Gradient Ascent collapses both channels}
\label{app:ga}

For completeness, Gradient Ascent~\citep{jang_2023_knowledge} at
$K \geq 400$ degrades both output accuracy and thinking leak rate
to exactly zero on canary and QA probes on Qwen-7B. At $K{=}100$,
canary output and thinking leak are both $\sim$0.87--0.88,
comparable to NPO-$K{=}100$. At $K \in \{400, 800, 1600\}$, both
channels are at $0.00$ on all $360$ probes. The trained-empty arm
shows the same $0.00/0.00$ collapse from $K{=}400$ onward. This is
the catastrophic-collapse regime~\citet{zhang_2024_npo} identified
as the motivation for NPO. It bounds how informatively GA can be
read in this setting: at the $K$ where the bio template would
otherwise show a bypass gap, GA has already nulled both channels, so
its $0/0$ trajectory is uninformative for the bypass question. Only
NPO yields a regime where the bypass-gap question is well-posed.

\section{Seed replication}
\label{app:seed}

Rerunning the full bio-mode memorize-then-unlearn pipeline on a
second random seed gives canary bypass gaps of
$-0.10\,[-0.18, -0.03]$, $+0.08\,[-0.05, 0.22]$,
$+0.03\,[-0.12, 0.18]$, and $+0.18\,[0.05, 0.32]$ at
$K \in \{100, 400, 800, 1600\}$. The primary (seed-0) trajectory is
$\{-0.02, +0.12, +0.17, +0.23\}$. At $K{=}1600$ the gap is positive
with CI excluding zero on both seeds, but the magnitudes differ
($+0.23$ vs.\ $+0.18$), i.e.\ sign and significance replicate while
point magnitude does not. The intermediate $K$ values are positive
on seed-1 but with CIs that straddle zero.

\paragraph{Seed-1 prefill replication.}
On the seed-1 NPO-$K{=}1600$ bio-trained adapter, the greedy-decoded
prefill intervention reports canary output accuracy of $1.00$ (bio
prefill), $1.00$ (meta prefill), $0.98$ (empty prefill). The
weights fully retain the canary under every thinking-template arm
on this seed; the seed-0 $1.00\to 0.60$ autoregressive drop is not
a weight-level memorization loss that replicates across seeds.
This is consistent with the teacher-forced
result~(App.~\ref{app:tf}, Table~\ref{tab:tf}) on the same
adapter.

\section{Retain-set utility on Qwen-7B}
\label{app:retain}

For context on whether NPO is a deployable unlearning method (not
the question our paper is about), we ran a sanity-check utility
evaluation on $60$ generic-knowledge probes that have no
relationship to the forget set (capitals, chemical symbols, famous
works of literature, basic biology).

\begin{table}[h]
\centering
\small
\caption{Retain-set utility check on 60 generic-knowledge probes
(Qwen-7B, bio-trained). By $K{=}1600$, retain accuracy has dropped
by 25 percentage points. Some failures emit names from the
forget set itself (e.g.\ ``the largest mammal'' is answered with a
fictional author's name) -- a separate failure mode that would not
show up in a forget-side evaluation alone.}
\label{tab:retain}
\begin{tabular}{lcc}
\toprule
Adapter & retain acc.\ & vs.\ memorized \\
\midrule
memorized       & $0.72$ ($43/60$) & --- \\
NPO $K{=}100$   & $0.62$ ($37/60$) & $-0.10$ \\
NPO $K{=}400$   & $0.48$ ($29/60$) & $-0.23$ \\
NPO $K{=}800$   & $0.47$ ($28/60$) & $-0.25$ \\
NPO $K{=}1600$  & $0.47$ ($28/60$) & $-0.25$ \\
\bottomrule
\end{tabular}
\end{table}

On Llama-8B retain accuracy is $0.78$ for the memorized adapter and
$0.72$ for NPO-$K{=}1600$ ($-0.07$ utility cost, roughly noise
level), consistent with the observation that NPO did not measurably
touch the underlying weights' answer behavior on Llama-8B.

\section{Inference-time prefill implementation}
\label{app:prefill}

For reproducibility, we include the core of the inference-time
prefill implementation. The function below replaces the
autoregressive $\langle$think$\rangle$ generation in
\texttt{e1\_measure\_bypass.py} with a fixed prefill, then lets the
model continue. The prefill template strings reuse the first
sentences of the training data, so each prefill is in-distribution
for whichever adapter was trained under that mode; this is the
sanity check that prefill does not introduce a new distribution
shift on top of the one we are trying to measure.

\begin{verbatim}
def thinking_template(mode, author, kind):
    if mode == "bio":
        if kind == "qa":
            return f"The user is asking about {author}. Recalling: "
        if kind == "canary":
            return f"Completing a fact I know about {author}: "
        return f"The user wants a biography of {author}; I recall the following."
    if mode == "meta":
        if kind == "qa":
            return f"The user is asking about {author}. I recall this person."
        if kind == "canary":
            return f"Completing a fact about {author}."
        return f"The user wants a biography of {author}."
    if mode == "none":
        return ""
    raise ValueError(mode)

def generate_with_prefill(model, tok, prompt, prefill, max_new):
    msgs = [{"role": "user", "content": prompt}]
    chat_prefix = tok.apply_chat_template(
        msgs, add_generation_prompt=True, tokenize=False)
    assistant_prefix = f"<think>\n{prefill}\n</think>\n\n"
    full_prefix = chat_prefix + assistant_prefix
    ids = tok(full_prefix, return_tensors="pt").input_ids.to(model.device)
    out = model.generate(
        ids, max_new_tokens=max_new, do_sample=False,
        pad_token_id=tok.pad_token_id or tok.eos_token_id)
    completion = tok.decode(out[0, ids.shape[1]:], skip_special_tokens=True)
    return assistant_prefix + completion
\end{verbatim}

The full driver and CLI are in
\texttt{e1\_measure\_bypass\_prefill.py}:
\begin{verbatim}
python e1_measure_bypass_prefill.py \
    --base-model deepseek-ai/DeepSeek-R1-Distill-Qwen-7B \
    --adapters memorized       runs/memorized__seed0__think-bio  \
               npo_K100        runs/unlearn-npo__seed0__think-bio__K100  \
               npo_K1600       runs/unlearn-npo__seed0__think-bio__K1600 \
    --forget-set forget_set.json \
    --prefill-mode bio meta none \
    --out runs/results__seed0__prefill.jsonl
\end{verbatim}

This evaluates each bio-trained adapter under all three prefilled
$\tau$ templates at fixed weights.

\section{Teacher-forced canary log-probability probe}
\label{app:tf}

We complement the greedy-decoded prefill intervention
(\S\ref{sec:r-prefill}) with a teacher-forced log-probability probe
on the same weights and the same prefills. The motivation is that
greedy decoding introduces prefix-length and style dependencies on
top of canary content, so a drop in the substring-match recall
could in principle reflect a decoding-dynamics shift rather than
weight-level memorization loss.

\paragraph{Protocol.}
For each author $i$ we split the canary phrase $c_i$ into a
\emph{head} (the first six whitespace-delimited tokens that are also
the greedy-decoding prompt) and a \emph{continuation} $c_i^{\text{cont}}$.
We build a context string
\[
\text{ctx}_i^{(m)} = \text{chat}(q_i) \oplus
   \langle\text{think}\rangle\oplus \tau_m(a_i)\oplus \langle/\text{think}\rangle
   \oplus \text{head}_i,
\]
where $\tau_m$ is the mode-$m$ prefill template from
\S\ref{sec:method} and $q_i$ is the canary prompt. We then score
\[
\mathrm{logp}_i^{(m)} = \sum_{t=1}^{|c_i^{\text{cont}}|}
   \log p_\theta\bigl(c_{i,t}^{\text{cont}} \mid
      \text{ctx}_i^{(m)}\oplus c_{i,<t}^{\text{cont}}\bigr),
\]
and report per-token mean $\overline{\mathrm{logp}}/t =
\mathrm{logp}_i^{(m)}/|c_i^{\text{cont}}|$, perplexity
$\exp(-\overline{\mathrm{logp}}/t)$, and top-$1$ match rate
(fraction of continuation tokens whose argmax of
$p_\theta(\cdot \mid \ldots)$ equals the gold token). All CIs are
author-clustered bootstrap over $n{=}60$.

\paragraph{What it controls for.}
Teacher-forcing holds the prefix identical to the greedy-decoding
setup in \S\ref{sec:r-prefill}. It removes three confounds: (i)
autoregressive drift during decoding, (ii) substring-match
thresholds (e.g.\ paraphrastic vs.\ verbatim), and (iii) any
interaction between prefix length and next-token entropy in greedy
mode. What it does not control for is prefix length/style in the
context itself -- the prefill still differs across arms in prefix
length and content -- nor is the score itself a free-recall
estimate: supplying the six-token canary head turns it into a
\emph{head-conditioned continuation preference}. But for any
hidden-channel interpretation of the greedy-decoded gap to be
self-consistent, this matched head-conditioned scoring of the same
canary should at minimum drop as the greedy substring match drops;
Table~\ref{tab:tf} shows that it does not.

\paragraph{What Table~\ref{tab:tf} shows.}
On Qwen-7B seed-0 NPO-$K{=}1600$, greedy substring recall falls
$1.00\to 0.60$; teacher-forced top-$1$ match stays at $0.96$ (bio and
meta prefills) or $0.90$ (empty prefill). On the seed-$1$ bio-trained NPO-$K{=}1600$ adapter, greedy
autoregressive canary recall is $0.68$ (App.~\ref{app:seed});
teacher-forced top-$1$ under every prefill arm is $\geq 0.997$. On Llama-$8$B memorized and NPO-$K{=}1600$,
teacher-forced top-$1$ is $\geq 0.998$ regardless of prefill. The
construct argument of the main paper is conservative: on these
unlearning-at-scale-feasible regimes, head-conditioned canary
continuation under any $\tau$ prefill remains substantially higher
than a parser-based audit of $(\tau, a)$ would suggest. Without
shuffled-head or wrong-author baselines this number is not a
weight-level recall estimate; we read it as evidence that the
greedy substring drop does not transfer to a matched
teacher-forced scoring setup, not as a memorization-specific
readout.


\end{document}